\documentclass[letterpaper, 10 pt, conference]{ieeeconf}  % Comment this line out
\IEEEoverridecommandlockouts                              % This command is only
\usepackage{graphics} % for pdf, bitmapped graphics files
\usepackage{booktabs}
\usepackage{epsfig} % for postscript graphics files
\usepackage{mathptmx} % assumes new font selection scheme installed
\usepackage{times} % assumes new font selection scheme installed
\usepackage{amsmath} % assumes amsmath package installed
\usepackage{amssymb}  % assumes amsmath package installed
\usepackage{tikz}
\usepackage{standalone}
\usepackage{bondgraphs}
\usepackage{subcaption}
\usepackage{url}
\usepackage{blox}
\usepackage{gensymb}
\usepackage{breakurl}
\usepackage{booktabs}
\usepackage{tabularx}
\usepackage{booktabs,amsfonts,dcolumn}
\usepackage{blindtext}
\usepackage{rotating}
\usetikzlibrary{shapes,arrows}
\usetikzlibrary{calc, patterns, decorations.pathmorphing, decorations.markings, shapes.geometric, arrows, positioning, angles}

\mathchardef\mhyphen="2D
\mathchardef\mslash="202F

\renewcommand{\arraystretch}{.8}
\usepackage{textcomp}
\usepackage{hyperref}
\usepackage{lipsum}
\allowdisplaybreaks

\title{\LARGE \bf
Complex Stiffness Model of Physical Human-Robot Interaction: Implications for Control of Performance Augmentation Exoskeletons}

\author{Binghan He$^{1}$, Huang Huang, Gray C. Thomas and Luis Sentis% <-this % stops a space
\thanks{This work was supported by the U.S. Government and NASA Space Technology Research Fellowship NNX15AQ33H. We would also like to thank the members of the Human Centered Robotics Lab, University of Texas at Austin and Apptronik Systems Inc for their supports. Authors are with The Departments of Mechanical Engineering (B.H., H.H., G.C.T.) and Aerospace Engineering (L.S.), University of Texas at Austin, Austin, TX. 
Send correspondence to $^{1}$$\;${\tt\small binghan at utexas dot edu}.
}
}

\newcommand\copyrighttext{%
  \scriptsize 
  Accepted for publication in IEEE/RSJ International Conference on Intelligent Robots and Systems (IROS)
  \textcopyright 2019 IEEE. Personal use of this material is permitted. Permission from IEEE must be obtained for all other uses, in any current or future media, including reprinting/republishing this material for advertising or promotional purposes, creating new collective works, for resale or redistribution to servers or lists, or reuse of any copyrighted component of this work in other works.
  DOI: \href{https://ieeexplore.ieee.org/document/8968005}{10.1109/IROS40897.2019.8968005}
}
\newcommand\copyrightnotice{%
\begin{tikzpicture}[remember picture,overlay]
\node[anchor=south,yshift=10pt] at (current page.south)
{\fbox{\parbox{\dimexpr\textwidth-\fboxsep-\fboxrule\relax}{\copyrighttext}}};
\end{tikzpicture}%
}

\usepackage{dcolumn}
\usepackage{caption}
\usepackage{multirow}
\begin{document}
\newcolumntype{L}[1]{>{\raggedright\arraybackslash}p{#1}}
\newcolumntype{C}[1]{>{\centering\arraybackslash}p{#1}}
\newcolumntype{R}[1]{>{\raggedleft\arraybackslash}p{#1}}

\maketitle
\thispagestyle{empty}
\pagestyle{empty}
\copyrightnotice

%%%%%%%%%%%%%%%%%%%%%%%%%%%%%%%%%%%%%%%%%%%%%%%%%%%%%%%%%%%%%%%%%%%%%%%%%%%%%%%%
\begin{abstract}
Human joint dynamic stiffness plays an important role in the stability of performance augmentation exoskeletons. In this paper, we consider a new frequency domain model of the human joint dynamics which features a complex value stiffness. This complex stiffness consists of a real stiffness and a hysteretic damping. We use it to explain the dynamic behaviors of the human connected to the exoskeleton, in particular the observed non-zero low frequency phase shift and the near constant damping ratio of the resonance as stiffness and inertia vary. We validate this concept with an elbow-joint exoskeleton testbed (attached to a subject) by experimentally varying joint stiffness behavior, exoskeleton inertia, and the strength augmentation gain. We compare three different models of elbow-joint dynamic stiffness: a model with real stiffness, viscous damping and inertia; a model with complex stiffness and inertia; and a model combining the previous two models. Our results show that the hysteretic damping term improves modeling accuracy (via a statistical F-test). Moreover, this term contributes more to model accuracy than the viscous damping term. In addition, we experimentally observe a linear relationship between the hysteretic damping and the real part of the stiffness which allows us to simplify the complex stiffness model down to a 1-parameter system. Ultimately, we design a fractional order controller to demonstrate how human hysteretic damping behavior can be exploited to improve strength amplification performance while maintaining stability.
\end{abstract}

%%%%%%%%%%%%%%%%%%%%%%%%%%%%%%%%%%%%%%%%%%%%%%%%%%%%%%%%%%%%%%%%%%%%%%%%%%%%%%%%
\section{Introduction}
While the concept of a personal augmentation device or exoskeleton is an old idea \cite{YagnNicolas1890PatentUSpatent, MakinsonBodineFitck1969report,KazerooniGuo1993JDSMC}, a system which delivers on the dream of transparent interaction, of ``feeling like the system is not there,'' through augmentation of sensed human interaction forces is still an ambitious goal of force control technology today \cite{Kazerooni2005IROS,DollarHerr2008TRO, JacobsenOlivier2014Patent, FontanaVertechyMarcheschiSalsedoBergamasco2014RAM}. Unlike assistive exoskeletons which help complete predictable behaviors \cite{ZhangFiersWitteJacksonPoggenseeAtkesonCollins2017Science,LeeKimBakerLongKaravasMenardGalianaWalsh2018JNR} or rehabilitation exoskeletons \cite{KongMoonJeonTomizuka2010TMech,KimDeshpande2017IJRR} which simulate rehabilitation therapy, human augmentation exoskeletons \cite{Kazerooni2005IROS, LeeLeeKimHanShinHan2014Mechatronics} use non-passive feedback control to amplify the user's strength. But this type of feedback control brings the system closer to instability. And since the exoskeleton is in a feedback interconnection with the human, a model of the human's dynamic behavior plays a critical role in determining the stability of an augmentation exoskeleton \cite{BuergerHogan2007TRO, HeThomasPaineSentis2019ACC}. 

Among all different kinds of dynamic model of an individual human joint, perhaps the most popular one is the mass-spring-damper model---with the additional non-linearity that the spring stiffness of the human joint can be modified by both voluntary muscle contractions or external torques exerted on the joint \cite{BennettHollerbachXuHunter1992EBR}. Several studies demonstrated a linear relationship between the stiffness of the human (found by fitting a linear mass-spring-damper model for a single joint) and an external torque \cite{AgarwalGottlieb1977JBE, CannonZahalak1982JB, HunterKearney1982JB}. For modeling the human joint damping, some other studies explored the fact that not only the stiffness but also the damping increases with muscle contractions \cite{BeckerMote1990JBE} and external torques \cite{WeissHunterKearney1988JB}. A linear relationship between the damping and the external torque has also been identified for the human ankle joint, but it is statistically weaker than the strong linear relationship between the stiffness of the ankle and the external torques \cite{AgarwalGottlieb1977JBE, HunterKearney1982JB}. However, it is not clear from the literature that a linear relationship between the damping and the stiffness of a human joint can be expected in more general cases. 

Another way to model the damping in the linear mass-spring-damper model is through the empirical observation that a relatively consistent damping ratio is maintained by the human elbow across different joint stiffnesses \cite{HeThomasPaineSentis2019ACC}. Frequency domain identification of the ankle joint impedance \cite{AgarwalGottlieb1977JBE, GottliebAgarwal1978JB} also showed a consistent damping ratio within the range from 0.22 to 0.49. This damping ratio consistency on the ankle is also supported by the fact that the ankle damping ratio does not have significant change with large variations of mean external torques exerted on the subjects \cite{WeissHunterKearney1988JB}. For upper limbs, a multi-joint impedance study on human arms \cite{PerreaultKirschCrago2004EBR} showed that the damping ratio of the minimally damped mode for the 2-D endpoint impedance in the transverse plane is distributed with a mean of 0.26 and a standard deviation of 0.08. Although this could be explained as the effect of humans adapting their damping to stabilize movement \cite{MilnerCloutier1993EBR}, a more detailed explanation of how humans achieve this consistency remains unclear.

Hysteretic damping models have seen success in biomechanical modelling before. In \cite{AgarwalGottlieb1977JBE}, experimental results showed a hysteretic relationship between the applied torque and the ankle angle at very low frequencies. Hysteretic damping is shown indirectly in \cite{CannonZahalak1982JB} (see Fig. 6 of that paper), where the human elbow stiffness has a phase shift around 25 degrees in a wide range of low frequencies---contradicting the viscous damping hypothesis. This type of phase behavior is explained (in the field of structural mechanics) by defining a hysteretic damping whose damping coefficient is proportional to the inverse of frequency \cite{BishopJohnson2011book}. Models with hysteretic damping have also been adapted to describe the dynamic properties of the whole body of a seated human \cite{KitazakiGriffin1997JB} as well as cockroach legs \cite{DudekFull2006JEB}.

In this paper we study the human stiffness and damping behavior when coupled to an exoskeleton inertia, and test the effectiveness of a hysteretic damping term in the system model. More specifically we compare three models 1) a linear mass, spring, and viscous damper model, 2) a nonlinear complex-stiffness-spring and mass model (that is, a spring, mass, and hysteretic damper model), and 3) a combination model with mass, spring, and both viscous and hysteretic damping. Our results show that there is a statistically significant benefit of the hysteretic damping term (comparing model 1 to model 3 with an F-test), and a less significant benefit for the viscous damping term (comparing model 2 to model 3). This hysteretic damping explains the consistent damping-ratio of the human--exoskeleton resonant peak even as the stiffness and exoskeleton inertia change---which is not well explained by the linear model. And it also explains the low frequency phase lag in human stiffness (previously observed in 
\cite{CannonZahalak1982JB}). Our elbow joint experiments vary parameters which would result in a differing damping ratio if the linear model were true: we change the inertia of the exoskeleton, and (indirectly, using an adjustable exercise hand grip and a bias torque) the stiffness of the human joint. We also test different exoskeleton strength amplification factors, and it does not appear to elicit a different human behavior than when the inertia is simply reduced. One further contribution of the paper is the theorizing of an amplification controller which uses fractional order filtering to exploit the hysteretic damping of the human, offering improved performance over previous strategies.

\section{Methods}

\subsection{Apparatus}
\begin{figure} [!tbp]
    \centering
        \scalebox{.94}{
        \def\svgwidth{.5\textwidth}
    	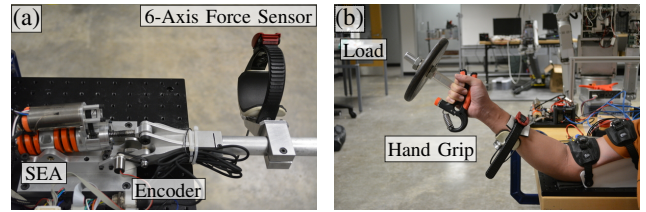}
    \caption{Experimental apparatus: a series elastic P0 exoskeleton from Apptronik Systems, featuring an ATI Mini40 force sensitive cuff and a P170 Orion air cooled series elastic actuator module acting through a simple 3 bar linkage.}
    \label{setup}
\end{figure}

For this study we employed the P0 series elastic elbow-joint exoskeleton from Apptronik Systems, as shown in Fig.~\ref{setup}. This exoskeleton has a moment of inertia of 0.1 $\rm{kg \cdot m^2}$ with no load on it, but allows for attaching additional weights to it. A load, attached 0.45 m from the exoskeleton joint, is pictured in Fig.~\ref{setup}.b. The contact force $f_c$ between the human and the exoskeleton is measured by a six-axis force/torque sensor situated below the white 3D printed ``cuff'' (which includes the adjustable strap which clamps the forearm). This force torque signal is cast as a torque ($\tau_c$) using the motion Jacobian $J$ of the sensor frame ($\tau_c = J^T f_c$). Rubber pads are adhered to the inside surfaces of the cuff and the cuff strap to improve user comfort. Joint position $\theta_e$ is directly measured by a dedicated encoder at the exoskeleton joint. The series elastic actuator (SEA) has a spring force control bandwidth of 10 Hz and provides high fidelity actuator torque $\tau_s$ tracking using the force control disturbance observer of \cite{PaineOhSentis2014TMech}. 

In parallel with an excitation chirp command (which essentially performs system identification of the human subject), a gravity compensation controller, a human augmentation controller, and a bias torque comprise the desired actuator torque signal. The gravity compensation controller takes the measurement of $\theta_e$ to calculate and compensate the gravity torque $\tau_g$ acting on the exoskeleton system. The human augmentation controller takes the measurement of $\tau_c$ and multiplies $\tau_c$ by negative $\alpha-1$. With the assistance of actuator torques produced from the augmentation command, the human's interaction forces with the exoskeleton are amplified by a factor of $\alpha$. This exoskeleton augmentation strategy differs from the one we applied in \cite{HeThomasPaineSentis2019ACC} in the directness of the augmentation feedback.

\subsection{Experimental Protocol}

The experimental protocol was approved by the Institutional Review Board (IRB) at the University of Texas at Austin. It consists of fifteen perturbation experiments with a 28-year old male subject. The experiments are separated into three groups (Exp. I-III) of five experiments. The first three experiments in each group are conducted with loads of 0.6 kg, 2.3 kg and 4.5 kg and an $\alpha$ value of 1 (corresponding to a non-augmentation controller) while the last two experiment in each group are conducted with a load of 4.5 kg and $\alpha$ values of 2 and 4. The mass of the loads and the mass of the exoskeleton have their gravitational bias torque fully compensated through gravity compensation control, while their inertia is attenuated by a factor of $\alpha$ due to the cuff torque feedback.

\begin{figure}[!tbp]
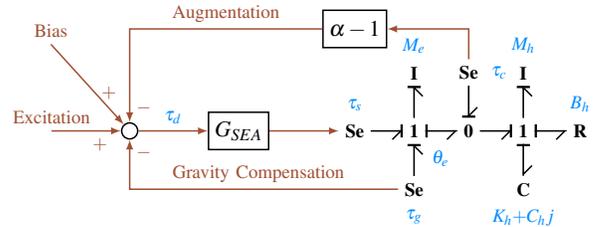

  	\centering
  	\scalebox{0.94}{
    \includestandalone[width=.5\textwidth]{block-diagram}}
    \caption{Block diagram consisting of augmentation, gravity compensation and experiment perturbation. Dynamics of human with exoskeleton are expressed as a bond graph with effort source of $\tau_s$, $\tau_c$ and $\tau_g$.}
    \label{block-diagram}
\end{figure}

The stiffness of the human elbow is influenced by muscle co-contraction as well as by contraction to resist the bias torque. In order to obtain different values of elbow stiffness for the three experiment groups, both the bias torque component of the controller and the co-contraction are varied. The three groups have, respectively, $0$ Nm, $4 \alpha$ Nm, and $8 \alpha$ Nm of bias torque. Co-contraction is controlled by having the subject squeeze an adjustable force hand grip. The three groups have a 10-kg, a 14-kg, and a 27-kg gripping force respectively. The amplitude of the perturbation chirp signal is set to be $2 \alpha$ Nm.

To avoid fatigue of the subject, the duration of each perturbation experiment is set to be 100 seconds. The perturbation is set to be an exponential chirp signal, and the results are typically  analyzed in the frequency domain. To sufficiently capture the natural frequency for damping feature identification, we set different ranges of frequency for the chirp signal according to the stiffness values the subject achieved from the bias torque and the gripping force. Frequency ranges of 2-20 rad/s, 3-30 rad/s and 4-40 rad/s are set for the chirp signals for the three experiment groups. 

After the chirp perturbation experiments, we transfer the time domain data into the frequency domain and identify the dynamic stiffness model of the subject by linear regression. The parameters of the three experiment groups are summarized in Tab.~\ref{exp-group}.

\subsection{Models}
In our models, we define $K_h$ as the human elbow-joint real stiffness, $C_h$ as the human elbow-joint hysteretic damping, $B_h$ as the human elbow-joint viscous damping, $M_h$ as the moment of inertia of the human and $M_e$ as the moment of inertia of the exoskeleton. See list of symbols in Tab.~\ref{symbols}.

A passive linear model of human dynamic stiffness with viscous damping can be expressed as
\begin{equation} \label{Sh1}
S_h = \tau_{c}/\theta_{e} = M_{h} s^2 + B_{h} s + K_{h}.
\end{equation}
If we consider a human model with hysteretic damping (complex stiffness) we have a nonlinear model
\begin{equation} \label{Sh2}
S_h = \tau_{c}/\theta_{e} = M_{h} s^2 + C_{h} j + K_{h}.
\end{equation}
% Where the hysteretic damping $C_h j$ helps to create a non-zero phase shift.
And to generalize the two, we also consider a nonlinear model with both viscous and hysteretic damping
\begin{equation} \label{Sh3}
S_h = \tau_{c}/\theta_{e} = M_{h} s^2 + B_{h} s + C_{h} j + K_{h}.
\end{equation}

However, these models are difficult to identify from the experimental $\tau_c$ and $\theta_e$ values because the natural frequency of the human dynamic stiffness can easily go beyond the range of the frequency for the experiments. With the augmentation controller, the operator feels an attenuated inertia from the exoskeleton. Therefore, we added a nominal attenuated inertia of $M_e/\alpha$ to the frequency domain data of $\tau_{c}/\theta_{e}$ for the model identification. In essence, we desensitize our identification to errors far above the natural frequency of the human spring and the exoskeleton inertia. Combining this additional term with \eqref{Sh1}, \eqref{Sh2} and \eqref{Sh3}, the three models of human-exoskeleton interaction can be expressed as
% \begin{equation}
\begin{align}
& S_{h \mhyphen e \mslash \alpha} = M_{h \mhyphen e \mslash \alpha} s^2 + B_{h} s + K_{h},  \label{M1} \tag{M1}  \\ 
& S_{h \mhyphen e \mslash \alpha} = M_{h \mhyphen e \mslash \alpha} s^2 + C_{h} j + K_{h},  \label{M2} \tag{M2}  \\
& S_{h \mhyphen e \mslash \alpha} = M_{h \mhyphen e \mslash \alpha} s^2 + B_{h} s + C_{h} j + K_{h},  \label{M3} \tag{M3}  
\end{align}
where $M_{h \mhyphen e \mslash \alpha} = M_h + M_e / \alpha$ is the perceived inertia at the human joint.

\setlength{\tabcolsep}{12.pt}
\renewcommand{\arraystretch}{1.2}

\begin{table} [!tbp]
\caption{Experiment Parameters}
\centering
\scalebox{1.}{
\begin{tabular}{L{0.2cm} C{0.1cm} C{0.2cm} C{0.2cm} C{0.2cm} C{.5cm} C{.8cm}} \toprule
Exp & $\alpha$ & Load (kg) & Grip (kg) & Bias (Nm) & Amplitude (Nm) &  Frequency (rad/s)  \\ [.5ex] 
 \midrule
 I.1 & $1$ & $0.6$ & \multirow{5}{*}{$10$} & \multirow{5}{*}{$0$} & \multirow{5}{*}{$2 \alpha$} & \multirow{5}{*}{$2-20$} \\
 I.2 & $1$ & $2.3$ &   &   &   &   \\
 I.3 & $1$ & $4.5$ &   &   &   &   \\
 I.4 & $2$ & $4.5$ &   &   &   &   \\
 I.5 & $4$ & $4.5$ &   &   &   &   \\ \midrule
 II.1 & $1$ & $0.6$ & \multirow{5}{*}{$14$} & \multirow{5}{*}{$4 \alpha$} & \multirow{5}{*}{$2 \alpha$} & \multirow{5}{*}{$3-30$} \\
 II.2 & $1$ & $2.3$ &   &   &   & \\
 II.3 & $1$ & $4.5$ &   &   &   & \\
 II.4 & $2$ & $4.5$ &   &   &   & \\
 II.5 & $4$ & $4.5$ &   &   &   & \\ \midrule
 III.1 & $1$ & $0.6$ & \multirow{5}{*}{$27$} & \multirow{5}{*}{$8 \alpha$} & \multirow{5}{*}{$2 \alpha$} & \multirow{5}{*}{$4-40$} \\
 III.2 & $1$ & $2.3$  &   &   &   & \\
 III.3 & $1$ & $4.5$  &   &   &   & \\
 III.4 & $2$ & $4.5$  &   &   &   & \\
 III.5 & $4$ & $4.5$  &   &   &   & \\ [0ex]
\bottomrule
\end{tabular}} \label{exp-group}
\end{table}

\setlength{\tabcolsep}{0.pt}
\renewcommand{\arraystretch}{1.2}

\begin{table} [!tbp]
\caption{List of Symbols}
\centering
\scalebox{.92}{
\begin{tabular}{L{1.6cm} L{7.2cm}} \toprule
Symbol & Meaning \\ [.5ex] 
 \midrule
$\tau_d$ & Actuator desired torque \\
$\tau_s$ & Actuator actual torque \\
$G_{SEA}$ & Transfer function from $\tau_d$ to $\tau_s$ \\
$\tau_c$ & Human-exoskeleton interaction torque \\
$\tau_g$ & Exoskeleton gravity torque \\
$\theta_e$ & Joint angular position \\
$K_h$ & Human joint real stiffness parameter\\
$C_h$ & Human joint hysteretic damping parameter \\
$B_h$ & Human joint viscous damping parameter \\
$M_h$ & Moment of inertia of human forearm \\
$M_e$ &  Moment of inertia of exoskeleton \\
$M_{h \mhyphen e}$ & Moment of inertia of human with exoskeleton \\
$M_{h \mhyphen e \mslash \alpha}$ & Moment of inertia of human with attenuated exoskeleton \\ 
$S_h$ & Joint dynamic stiffness of human \\
$S_{h \mhyphen e}$ & Joint dynamic stiffness of human with exoskeleton \\
$S_{h \mhyphen e \mslash \alpha}$ & Joint dynamic stiffness of human with attenuated exoskeleton \\ 
$P_{\alpha}, \, C_{\alpha}$ & Plant and controller transfer functions of augmentation  \\
$\omega_{SEA}$ & Natural frequency of $G_{SEA}$ \\ 
$\omega_h, \, \zeta_h$ & Natural frequency and damping ratio of $S_h$ \\
$\omega_{h \mhyphen e}, \, \zeta_{h \mhyphen e}$ & Natural frequency and damping ratio of $S_{h \mhyphen e}$ \\ 
$\omega_{h \mhyphen e \mslash \alpha}, \, \zeta_{h \mhyphen e \mslash \alpha}$ & Natural frequency and damping ratio of $S_{h \mhyphen e \mslash \alpha}$ \\ [0ex]
\bottomrule
\end{tabular}} \label{symbols}
\end{table}

\begin{figure*}[!tbp]
    \centering
        \scalebox{1.}{
        \def\svgwidth{1.\textwidth}
    	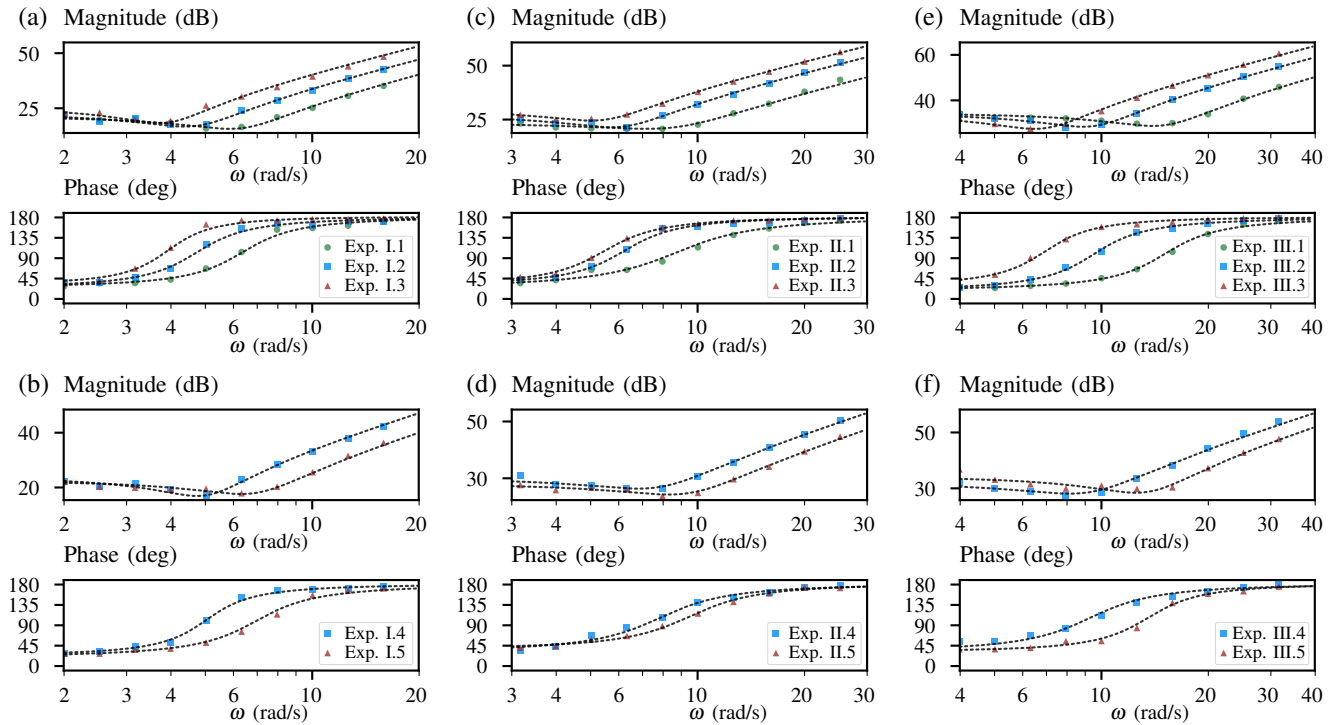}
    \caption{Bode plots of frequency domain data of $S_{h \mhyphen e \mslash \alpha}$ with Exp. I.1-5 on (a) and (b), Exp. II.1-5 on (c) and (d), and Exp. III.1-5 on (e) and (f). The dash lines on each plot show the fitted curves from M3.}
    \label{bode}
\end{figure*}

%To understand the stability of the human controlling the perceived inertia with the dynamic stiffness, 
We also calculate the damping ratio $\zeta_{h \mhyphen e \mslash \alpha}$ of $S_{h \mhyphen e \mslash \alpha}$, as a measure of the degree of oscillation at the resonant zero-pair. Because M2 and M3 have the $C_h j$ term which provides a damping effect in addition to $B_h s$, we define the damping ratio of each model using the imaginary part of the transfer function evaluated at the resonance:
\begin{flalign}
& \zeta_{h \mhyphen e \mslash \alpha} = \frac{B_h}{2 \sqrt{K_h M_{h \mhyphen e \mslash \alpha}}} \quad \text{for M1,} \label{ztm1} \\
& \zeta_{h \mhyphen e \mslash \alpha} = \frac{C_h \omega_{h \mhyphen e \mslash \alpha}^{-1}}{2 \sqrt{K_h M_{h \mhyphen e \mslash \alpha}}} = \frac{C_h}{2 K_h} \quad \text{for M2, and} \label{ztm2} \\ 
& \zeta_{h \mhyphen e \mslash \alpha} = 
%\frac{B_h + C_h \omega_{h \mhyphen e \mslash \alpha}^{-1}}{2 \sqrt{K_h M_{h \mhyphen e \mslash \alpha}}} =
\frac{B_h}{2 \sqrt{K_h M_{h \mhyphen e \mslash \alpha}}} + \frac{C_h}{2 K_h} \quad \text{for M3,} \label{ztm3}
\end{flalign}
where \begin{small} $\omega_{h \mhyphen e \mslash \alpha} = \sqrt{K_h/M_{h \mhyphen e \mslash \alpha}}$ \end{small} is the natural frequency of $S_{h \mhyphen e \mslash \alpha}$.

\subsection{Statistical Analysis}
In order to compare the significance of $B_h s$ and $C_h j$ in the human-exoskeleton interaction model, we calculate the residual square sum (RSS) for all three models, denoted $RSS_{M1}$, $RSS_{M2}$ and $RSS_{M3}$ respectively. For each experiment, we conduct F-tests for each of the two three-parameter models (M1 and M2) against the generalizing four-parameter model (M3). Our F-statistic accounts for complex value data,
\begin{flalign}
& F_{Mi \mhyphen M3} = \frac{RSS_{Mi} - RSS_{M3}}{RSS_{M3}} (2n - 4), \quad \text{for} \; i = 1, \, 2
\end{flalign}
where $n$ is the number of complex value samples at the frequency domain and the real and imaginary parts of each sample are statistically independent. The significance of $B_h s$ and $C_h j$ then will be evaluated by comparing this F statistic against a critical F statistic threshold based on a 0.05 false-rejection probability.

We split the 100 seconds of time domain data for each experiment into 10 sequences. For each of the 10 second sequences, only the data from the first 5.78 seconds is used for calculating the frequency domain sample. The remainder period of 4.22 seconds is greater than the 2\% settling time for all the 2nd order dynamics of $S_{h \mhyphen e \mslash \alpha}$ identified in the experiments. By this method we can safely assume statistical independence between the 10 single-frequency data points comprising our estimate of the frequency response function for the purposes of statistical testing.

\section{Results}

\subsection{Phase Shift}

In the frequency domain results of $S_{h \mhyphen e \mslash \alpha}$ (Fig.~\ref{bode}), the phase starts (at low frequencies) from a value between $25\degree$ to $45\degree$ instead of zero and changes very little across all the frequencies before it reaches the second order zero at $\omega_{h \mhyphen e \mslash \alpha}$ for each experiment. This type of phase shift is very different from the phase shift usually experienced by a linear system with a constant time delay or a constant damping property in which the phase shift approaches zero in the limit as $\omega\rightarrow0$. As shown in Fig.~\ref{time}, this phase shift is clearly visible even in time domain comparisons of $\tau_c$ and $\theta_e$. The data show that the human joint motion $\theta_e$ is not perfectly sinusoidal---it stops following the trend of the torque after they both reach their peak values and ``waits'' before following the torque $\tau_c$ in its descent. At low frequencies, these peaks seem especially flat.

\subsection{Model Comparisons}

The results of the identified parameters (Tab.~\ref{parameter}) show that the three models give the same values of $K_h$, $M_{h \mhyphen e \mslash \alpha}$ and consequently $\omega_{h \mhyphen e \mslash \alpha}$ to two decimal places for each experiment. This is because the difference between the three models is restricted to the imaginary part of $S_{h \mhyphen e \mslash \alpha}$ while $K_h$ and $M_{h \mhyphen e \mslash \alpha}$ are the coefficients of the real part of $S_{h \mhyphen e \mslash \alpha}$. Although the identified values of $B_h$ and $C_h$ are quite different between the three models, the values of $\zeta_{h \mhyphen e \mslash \alpha}$ are still very close for each experiment. This means that the three models give very similar values for the slope of the phase at the resonant frequency $\omega_{h \mhyphen e \mslash \alpha}$.

From M1 to M3, the values of $B_h$ have been reduced considerably. This means that M3 uses the $C_h j$ term to replace part of the $B_h$ term in M1 while maintaining a similar phase behavior at the frequency $\omega_{h \mhyphen e \mslash \alpha}$. From M2 to M3, the values of $C_h$ have been reduced except for Exp. I.3, III.3 and III.5 in which M3 gives a negative value for $B_h$. These negative value of $B_h$ is because there is no lower bound constraint on the value of $B_h$ during the frequency domain regression for M3. Although a negative value of $B_h$ brings non-passivity to a linear mass-spring-damper system in the common sense, the $C_h j$ term in M3 enforces the dynamics of $S_{h \mhyphen e \mslash \alpha}$ to remain passive across the range of frequencies in our experiments. 

The results from the F-tests (Fig.~\ref{f-statistics}) relate to the significance of $B_h s$ and $C_h j$ in M3. Based on the 20 statistically independent data values for each experiment, a critical F-statistic value of 4.49 is calculated for 0.05 false-rejection probability. The results show that values of $F_{M1 \mhyphen M3}$ for all the experiments are much higher than the critical F-statistic value, with the values of $F_{M1 \mhyphen M3}$ in Exp. II.3 and II.5 exceeding 100 (c.f. the critical value of 4.49). This proves that the existence of the $C_h j$ term in M3 significantly improves modeling accuracy of $S_{h \mhyphen e \mslash \alpha}$. The values of $F_{M2 \mhyphen M3}$ are mostly below the critical F-statistic value except for Exp. I.5, II.1, III.1 and III.2. The other observation is that the value of $F_{M2 \mhyphen M3}$ is always much lower than the value of $F_{M2 \mhyphen M3}$ for all experiments. Although the effect of the $B_h s$ term cannot be completely ignored based on the results of these F-tests, we can claim that the $C_h j$ term is still much more significant than the $B_h s$ term in M3.

\subsection{Linear Regression between $C_h$ and $K_h$}

Because the $C_h j$ term is created to describe the phase shift effect from the complex human stiffness in M2 and M3, we suspect that the identified value of $C_h$ has a linear relation with the value of $K_h$. Therefore, we apply linear regression between the values of $C_h$ and $K_h$ identified from M2 and M3 (Fig.~\ref{regression}). Compared with M3, the linear regression result with M2 shows a stronger linear relationship with a much higher coefficient of determination ($R^2$). The regression equation identified from the M2 parameters also has a smaller value of bias from the origin of the $C_h \mhyphen K_h$ plane compared with the regression equation identified from the M3 parameters. Intuition leads us to expect low bias in the regression equation, since a nonzero value of $C_h$ when the value of $K_h$ is zero could not be explained as hysteretic spring behavior.

\setlength{\tabcolsep}{3.pt}
\renewcommand{\arraystretch}{0.9}
\begin{table}
\caption{Subject Dynamic Stiffness Parameters}
\centering
\scalebox{.92}{
\begin{tabular}{l c c c c c c c} \toprule
\footnotesize Exp & \footnotesize Model & \footnotesize$K_h (\frac{N m}{rad})$ & \footnotesize$C_h (\frac{N m}{rad})$ & \footnotesize$B_h (\frac{N m s}{rad})$ & \footnotesize$M_{h \mhyphen e \mslash \alpha}$\scriptsize$(kg m ^ {2})$ & \footnotesize$\omega_{h \mhyphen e \mslash \alpha} (\frac{rad}{s})$ & \footnotesize$\zeta_{h \mhyphen e \mslash \alpha}$ \\ [.5ex] 
 \midrule
% Exp.I\mhyphen1   & $1.0$ & $10.05$ &  $4.97$ &  $0.18$ & $0.28$ & $5.95$  & $0.30$ \\
% Exp.I\mhyphen2   & $1.0$ & $11.80$ &  $5.44$ &  $0.31$ & $0.60$ & $4.44$  & $0.29$ \\

\multirow{3}{*}{I.1}   & M1 & $10.05$ & $\mhyphen \, \mhyphen$ & $1.03$     & $0.28$ &  $5.95$ &  $0.31$ \\
                            & M2 & $10.05$ & $ 5.89$    & $\mhyphen \, \mhyphen$ & $0.28$ &  $5.95$ &  $0.29$ \\
                            & M3 & $10.05$ & $ 4.97$    & $0.18$     & $0.28$ &  $5.95$ &  $0.30$ \\ 
\cmidrule[.5pt]{2-8} 
\multirow{3}{*}{I.2}   & M1 & $11.80$ & $\mhyphen \, \mhyphen$ & $1.51$     & $0.60$ &  $4.44$ &  $0.28$ \\
                            & M2 & $11.80$ & $ 6.68$    & $\mhyphen \, \mhyphen$ & $0.60$ &  $4.44$ &  $0.28$ \\
                            & M3 & $11.80$ & $ 5.44$    & $0.31$     & $0.60$ &  $4.44$ &  $0.29$ \\ 
\cmidrule[.5pt]{2-8} 
\multirow{3}{*}{I.3}   & M1 & $15.74$ & $\mhyphen \, \mhyphen$ & $2.09$     & $1.18$ &  $3.65$ &  $0.24$ \\
                            & M2 & $15.74$ & $ 8.33$    & $\mhyphen \, \mhyphen$ & $1.18$ &  $3.65$ &  $0.26$ \\
                            & M3 & $15.74$ & $10.44$    &$-0.60$     & $1.18$ &  $3.65$ &  $0.26$ \\ 
\cmidrule[.5pt]{2-8} 
\multirow{3}{*}{I.4}   & M1 & $13.82$ & $\mhyphen \, \mhyphen$ & $1.46$     & $0.60$ &  $4.78$ &  $0.25$ \\
                            & M2 & $13.82$ & $6.87$     & $\mhyphen \, \mhyphen$ & $0.60$ &  $4.78$ &  $0.25$ \\ 
                            & M3 & $13.82$ & $6.01$     & $0.21$     & $0.60$ &  $4.78$ &  $0.25$ \\ 
\cmidrule[.5pt]{2-8} 
\multirow{3}{*}{I.5}   & M1 & $12.09$ & $\mhyphen \, \mhyphen$ & $1.22$     & $0.28$ &  $6.59$ &  $0.33$ \\
                            & M2 & $12.09$ & $6.84$     & $\mhyphen \, \mhyphen$ & $0.28$ &  $6.59$ &  $0.28$ \\
                            & M3 & $12.09$ & $4.26$     & $0.52$     & $0.28$ &  $6.59$ &  $0.32$ \\
\midrule
\multirow{3}{*}{II.1}  & M1 & $12.73$ & $\mhyphen \, \mhyphen$ & $1.41$     & $0.20$ &  $7.94$ &  $0.44$ \\
                            & M2 & $12.73$ & $10.18$    & $\mhyphen \, \mhyphen$ & $0.20$ &  $7.94$ &  $0.40$ \\
                            & M3 & $12.73$ & $ 5.86$    & $0.66$     & $0.20$ &  $7.94$ &  $0.44$ \\
\cmidrule[.5pt]{2-8}
\multirow{3}{*}{II.2}  & M1 & $18.79$ & $\mhyphen \, \mhyphen$ & $1.91$     & $0.57$ &  $5.72$ &  $0.29$ \\
                            & M2 & $18.79$ & $11.77$    & $\mhyphen \, \mhyphen$ & $0.57$ &  $5.72$ &  $0.31$ \\
                            & M3 & $18.79$ & $11.54$    & $0.04$     & $0.57$ &  $5.72$ &  $0.31$ \\
\cmidrule[.5pt]{2-8}
\multirow{3}{*}{II.3}  & M1 & $25.95$ & $\mhyphen \, \mhyphen$ & $3.08$     & $1.03$ &  $5.02$ &  $0.30$ \\
                            & M2 & $25.95$ & $16.75$    & $\mhyphen \, \mhyphen$ & $1.03$ &  $5.02$ &  $0.32$ \\ 
                            & M3 & $25.95$ & $15.48$    & $0.26$     & $1.03$ &  $5.02$ &  $0.32$ \\
\cmidrule[.5pt]{2-8}
\multirow{3}{*}{II.4}  & M1 & $25.77$ & $\mhyphen \, \mhyphen$ & $2.83$     & $0.52$ &  $7.02$ &  $0.39$ \\
                            & M2 & $25.77$ & $20.49$    & $\mhyphen \, \mhyphen$ & $0.52$ &  $7.02$ &  $0.40$ \\
                            & M3 & $25.77$ & $16.60$    & $0.60$     & $0.52$ &  $7.02$ &  $0.40$ \\
\cmidrule[.5pt]{2-8}
\multirow{3}{*}{II.5}  & M1 & $19.07$ & $\mhyphen \, \mhyphen$ & $1.88$     & $0.28$ &  $8.32$ &  $0.41$ \\
                            & M2 & $19.07$ & $16.27$    & $\mhyphen \, \mhyphen$ & $0.28$ &  $8.32$ &  $0.43$ \\
                            & M3 & $19.07$ & $15.72$    & $0.08$     & $0.28$ &  $8.32$ &  $0.43$ \\
\midrule
\multirow{3}{*}{III.1} & M1 & $48.15$ & $\mhyphen \, \mhyphen$ &$ 1.97$     & $0.23$ & $ 14.4$ &  $0.29$ \\ 
                            & M2 & $48.15$ & $25.45$    &$\mhyphen \, \mhyphen$ & $0.23$ & $ 14.4$ &  $0.26$ \\
                            & M3 & $48.15$ & $16.66$    &$ 0.76$     & $0.23$ & $ 14.4$ &  $0.29$ \\
\cmidrule[.5pt]{2-8}
\multirow{3}{*}{III.2} & M1 & $48.60$ & $\mhyphen \, \mhyphen$ &$ 2.85$    & $0.58$ & $ 9.13$ &  $0.27$ \\
                            & M2 & $48.60$ & $25.61$    &$\mhyphen \, \mhyphen$ & $0.58$ & $ 9.13$ &  $0.26$ \\
                            & M3 & $48.60$ & $15.19$    &$ 1.23$    & $0.58$ & $ 9.13$ &  $0.27$ \\
\cmidrule[.5pt]{2-8}
\multirow{3}{*}{III.3} & M1 & $42.23$ & $\mhyphen \, \mhyphen$ &$ 3.19$    & $1.01$ & $ 6.47$ &  $0.24$ \\ 
                            & M2 & $42.23$ & $23.60$    &$\mhyphen \, \mhyphen$ & $1.01$ & $ 6.47$ &  $0.28$ \\
                            & M3 & $42.23$ & $24.08$    &$-0.07$    & $1.01$ & $ 6.47$ &  $0.28$ \\
\cmidrule[.5pt]{2-8}
\multirow{3}{*}{III.4} & M1 & $32.22$ & $\mhyphen \, \mhyphen$ &$ 2.82$    & $0.46$ & $ 8.35$ &  $0.37$ \\
                            & M2 & $32.22$ & $25.36$    &$\mhyphen \, \mhyphen$ & $0.46$ & $ 8.35$ &  $0.39$ \\
                            & M3 & $32.22$ & $20.83$    &$ 0.55$    & $0.46$ & $ 8.35$ &  $0.39$ \\
\cmidrule[.5pt]{2-8}
\multirow{3}{*}{III.5} & M1 & $42.33$ & $\mhyphen \, \mhyphen$ &$ 2.08$    & $0.27$ & $12.43$ &  $0.31$ \\
                            & M2 & $42.33$ & $26.50$    &$\mhyphen \, \mhyphen$ & $0.27$ & $12.43$ &  $0.31$ \\
                            & M3 & $42.33$ & $27.66$    &$-0.11$    & $0.27$ & $12.43$ &  $0.31$ \\ [0ex]
\bottomrule
\end{tabular}} \label{parameter}
\end{table}

\begin{figure}[tp!]
    \centering
        \scalebox{.90}{
        \def\svgwidth{.5\textwidth}
    	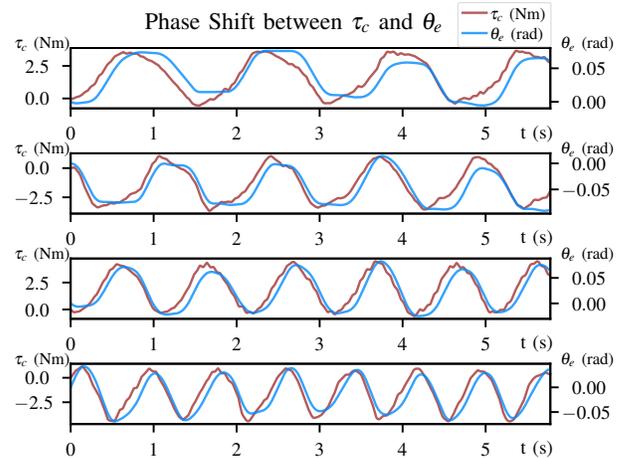}
    \caption{Four pieces of time data of $\tau_c$ and $\theta_e$ from Exp.III.1 used for identifying the frequency data of $S_{h \mhyphen e \mslash \alpha}$ at the frequencies of 4.0, 5.0, 6.3 and 8.0 rad/s show the phase shift in the time domain.}
    \label{time}
\end{figure}

Based on linear regression equations, we can express the phase shift (with respect to 0\degree) at the low frequencies as
\begin{flalign}
\text{\small Phase Shift} & = \tan^{-1}(\frac{C_h}{K_h}) = \tan^{-1}(c_h + \frac{d_h}{K_h}) \; \text{for M2, and} \label{phase-lag-2} \\ 
\text{\small Phase Shift} & =
%\tan^{-1}(\frac{C_h + B_h \omega}{K_h}) & \notag \\& =
\tan^{-1}(c_h + \frac{d_h + B_h \omega}{K_h}) \qquad \rm{for} \; M3,  \label{phase-lag-3}
\end{flalign}
where $C_h = c_h K_h + d_h$ is the regression equation identified from the values of $C_h$ and $K_h$ in M2 and M3 with $c_h$ and $d_h$ being the slope and the bias of the regression equation. By substituting $C_h = c_h K_h + d_h$ into \eqref{ztm2} and \eqref{ztm3}, the value of $\zeta_{h \mhyphen e \mslash \alpha}$ for M2 and M3 can be expressed as
\begin{flalign}
\zeta_{h \mhyphen e \mslash \alpha} & = \frac{C_h}{2 K_h} = \frac{c_h}{2} + \frac{d_h}{2 K_h} \qquad \rm{for} \; M2, \text{ and} \label{ztm2-new} \\ 
\zeta_{h \mhyphen e \mslash \alpha} & = 
%\frac{C_h}{2 K_h} + \frac{B_h}{2 \sqrt{K_h M_{h \mhyphen e \mslash \alpha}}} & \notag \\& = 
\frac{c_h}{2} + \frac{d_h}{2 K_h} + \frac{B_h}{2 \sqrt{K_h M_{h \mhyphen e \mslash \alpha}}} \qquad \rm{for} \; M3.  \label{ztm3-new}
\end{flalign}
Because the values of $d_h$ of the regression equations for M2 and M3 and the values of $B_h$ for M3 are relatively small, the phase shift at the low frequencies is dominated by the value of $\tan^{-1}(c_h)$ and the value of $\zeta_{h \mhyphen e \mslash \alpha}$ is dominated by the constant $c_h / 2$ term. This explains the fact that the phase shift is non-zero at low frequencies and the fact that the value of $\zeta_{h \mhyphen e \mslash \alpha}$ changes very little compared to the changes of $K_h$ and $M_{h \mhyphen e \mslash \alpha}$ across all our experiments.

\section{Implications for Control of Performance Augmentation Exoskeletons}

\subsection{1-Parameter Complex Stiffness Model}

One of the challenges of augmentation control is to design an augmentation controller to stabilize the exoskeleton with all possible human impedances. This requires a robust human impedance model with bounded parameter uncertainties for the augmentation controller design. 

Similar to \cite{BuergerHogan2007TRO}, a robust model version of M1 can be defined with bounded uncertainties for $K_h$ and $B_h$, which could be obtained from multiple measurements in advance. (We assume $M_h$ does not change for the elbow-joint.) Because both $K_h$ and $B_h$ vary in large ranges, the 2-D uncertain parameter space of $K_h \mhyphen B_h$ becomes very huge, and the augmentation controller can easily end up as an extremely low-bandwidth conservative controller. Since such an uncertain model includes all combinations of possible $K_h$ and $B_h$, the damping ratio can be a very limiting design constraint. This is not realistic, given that $\zeta_{h \mhyphen e \mslash \alpha}$ is relatively consistent in our experiment results. Therefore, we propose a 1-parameter model simplification to reduce the uncertain parameter space for augmentation controller design. 

One strategy is to model $B_h$ as a linear function of $K_h$ which allows us to create a robust model of M1 with bounded uncertainty for only $K_{h}$. Based on a linear relationship $B_h = a_h K_h$ between $B_h$ and $K_h$, M1 can be expressed as
\begin{equation} \label{M1-new-1}
S_{h \mhyphen e \mslash \alpha} = M_{h \mhyphen e \mslash \alpha} s^2 + K_{h}(1 + a_{h} s).
\end{equation}
By substituting $B_h = a_h K_h$ to \eqref{ztm1}, $\zeta_{h \mhyphen e \mslash \alpha}$ can be expressed as
\begin{flalign}
& \zeta_{h \mhyphen e \mslash \alpha} = \frac{a_h K_h}{2 \sqrt{K_h M_{h \mhyphen e \mslash \alpha}}} = \frac{a_h}{2} \cdot \omega_{h \mhyphen e \mslash \alpha}, \label{ztm1-new-1} 
\end{flalign}
which is proportional to $\omega_{h \mhyphen e \mslash \alpha}$. However, we do not observe this proportional relationship between $\zeta_{h \mhyphen e \mslash \alpha}$ and $\omega_{h \mhyphen e \mslash \alpha}$ from our experimental results for M1 in Tab.~\ref{parameter}. On the other hand, because \eqref{M1-new-1} is a simplification from M1, it also fails to explain the non-zero phase shift at low frequencies.

\begin{figure}[tp!]
    \centering
        \scalebox{.90}{
        \def\svgwidth{.5\textwidth}
    	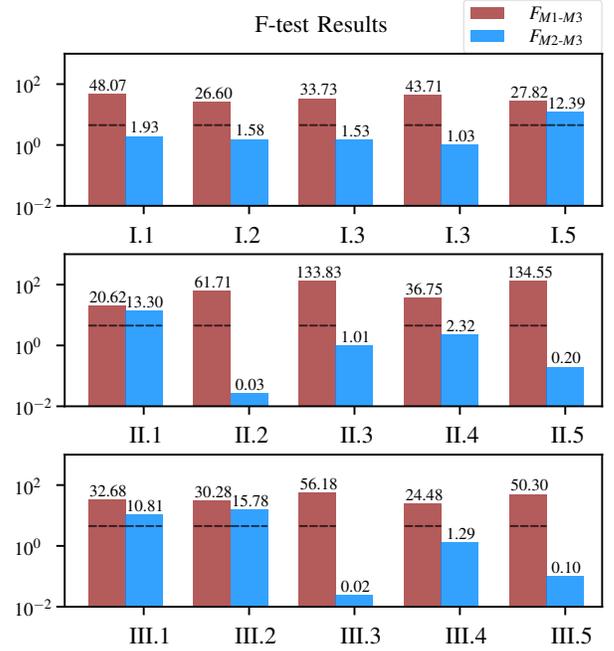}
    \caption{F-statistics on log scale for all experiments show the significant improvement on modeling accuracy from M1 to M3 and a partial improvement from M2 to M3. The dashed line appears on a bar if the F-statistic value is over the critical F-statistic value of 4.49 (false-rejection probability of 0.05).}
    \label{f-statistics}
\end{figure}

\begin{figure}
    \centering
        \scalebox{0.90}{
        \def\svgwidth{.5\textwidth}
    	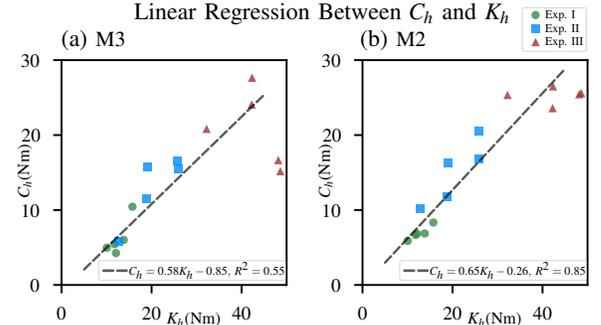}
    \caption{Linear regressions between $C_h$ and $K_h$ for M3 (a) and M2 (b) show that the parameters of M2 have a stronger linear relationship (that is, a higher $R^2$ value).}
    \label{regression}
\end{figure}

If we assume $d_h \approx 0$, a simplified complex stiffness model of M2 can be expressed as 
\begin{equation} \label{M2-new}
S_{h \mhyphen e \mslash \alpha} = M_{h \mhyphen e \mslash \alpha} s^2 + K_{h}(1 + c_{h} j).
\end{equation}
Based on \eqref{phase-lag-2} and \eqref{ztm2-new}, \eqref{M2-new} is able to explain both the non-zero phase shift at low frequencies and the near constant value of $\zeta_{h \mhyphen e \mslash \alpha}$ across all the experiments. This, in turn, supports the use of \eqref{M2-new} as a 1-parameter model of $S_{h \mhyphen e \mslash \alpha}$ for augmentation controller design. 

Adopting this 1-parameter model allows simplifying \eqref{Sh2},
\begin{equation} \label{Sh2-new}
S_{h} = \tau_{c}/\theta_{e} = M_{h} s^2 + K_{h}(1 + c_{h} j),
\end{equation}
and the dynamic stiffness of the human coupled with the exoskeleton $S_{h \mhyphen e}$,
\begin{equation} \label{Sh-e2-new}
S_{h \mhyphen e} = \tau_{s}/\theta_{e} = M_{h \mhyphen e} s^2 + K_{h}(1 + c_{h} j),
\end{equation}
where $M_{h \mhyphen e} = M_h + M_e$ is the combined inertia between the human and the exoskeleton. We consider $\omega_{h \mhyphen e} = \sqrt{K_h / M_{h \mhyphen e}}$ to be the natural frequency of $S_{h \mhyphen e}$, despite the $c_h$ term. 

\subsection{Fractional-Order Augmentation Controller}

As in \cite{HeThomasPaineSentis2019ACC}, the augmentation control we discuss here is designed to eliminate the augmentation error signal $\tau_{\alpha} = (\alpha - 1) \tau_c + \tau_s$ by feeding it back to the actuator command $\tau_d$ with an augmentation controller. Different from the direct augmentation feedback shown in Fig.~\ref{block-diagram} in which the augmentation command is $-\tau_c$ multiplied by $\alpha -1$, this strategy allows us to design an augmentation controller completely separated from the augmentation factor $\alpha$. 

By substituting \eqref{Sh2-new} and \eqref{Sh-e2-new}, the transfer function from $\tau_s$ to $\tau_{\alpha}$ can be expressed as
\begin{align} \label{a*Sh-e-a}
\frac{\tau_\alpha}{\tau_s} = \frac{(\alpha - 1) \cdot S_h + S_{h \mhyphen e}}{S_{h \mhyphen e}} = \alpha \cdot \frac{S_{h \mhyphen e \mslash \alpha}}{S_{h \mhyphen e}}.
\end{align}
Based on \eqref{a*Sh-e-a}, the augmentation plant transfer function $P_{\alpha}$ from $\tau_d$ to $\tau_{\alpha}$ can then be expressed as
\begin{equation}
P_{\alpha}(s) = \frac{\tau_\alpha}{\tau_d} = \alpha \cdot \frac{S_{h \mhyphen e \mslash \alpha}}{S_{h \mhyphen e}} \cdot G_{SEA} (s),
\end{equation}
where the SEA transfer function $G_{SEA} (s) = \tau_s / \tau_d$ acts as a 2nd order low-pass filter. Because of the high bandwidth of the SEA force controller, the natural frequency $\omega_{SEA}$ of $G_{SEA}$ is much greater than both $\omega_{h \mhyphen e}$ and $\omega_{h \mhyphen e \mslash \alpha}$.

\begin{figure}
    \centering
        \scalebox{0.90}{
        \def\svgwidth{.5\textwidth}
    	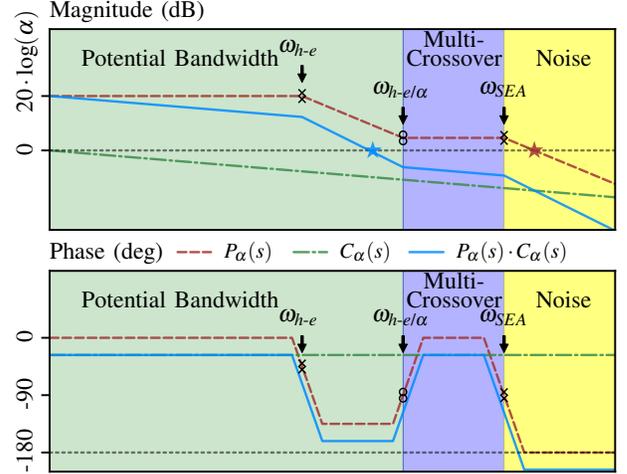}
    \caption{Conceptual bode plots show the augmentation plant $P_\alpha(s)$ with its poles (crosses) and zeros (circles). 
    Regions are color-coded: the model is trustworthy in the green region, the blue region reflects the multi-crossover behavior which makes an augmentation controller design unreliable, and the yellow region is dominated by sensor noise from $\tau_c$. 
    A fractional-order filter $C_\alpha(s)$ brings $P_\alpha(s)$ to a lower crossover and increases the phase margin. The stars indicate the crossovers of $P_\alpha(s)$ and $P_\alpha(s)$ with $C_\alpha(s)$.}
    \label{concept}
\end{figure}

Looking at the bode plot from low to high frequencies, $P_{\alpha} (s)$ has a pair of conjugate poles at $\omega_{h \mhyphen e}$, then a pair of conjugate zeros at $\omega_{h \mhyphen e \mslash \alpha}$ and then another pair of conjugate poles at $\omega_{SEA}$ (Fig.~\ref{concept}). Before $\omega_{h \mhyphen e}$, both $S_{h \mhyphen e}$ and $S_{h \mhyphen e \mslash \alpha}$ are dominated by the complex stiffness. Therefore, $P_{\alpha} (s)$ has magnitude $\alpha$ and $0^\circ$ phase. Between $\omega_{h \mhyphen e}$ and $\omega_{h \mhyphen e \mslash \alpha}$, $S_{h \mhyphen e}$ is dominated by its inertia effect and the magnitude of $P_{\alpha} (s)$ decreases while the phase leaves $0^\circ$. On the other hand, $S_{h \mhyphen e \mslash \alpha}$ is still dominated by the complex stiffness and prevents the phase moving below $\tan^{-1}(c_h) - 180\degree$. At the frequency between $\omega_{h \mhyphen e \mslash \alpha}$ and $\omega_{SEA}$, the inertia effects in $S_{h \mhyphen e}$ and $S_{h \mhyphen e \mslash \alpha}$ completely dominate their frequency behaviors. The magnitude of $P_{\alpha} (s)$ stays at $(\alpha M_h + M_e) / (M_h + M_e)$ which is in the range from 1 to $\alpha$. 

However, the gain crossover of $P_{\alpha}$ falls beyond $\omega_{SEA}$ without an augmentation controller. The phase margin with such crossover is very close to zero because of the 2nd order SEA dynamics. Also, the closed loop behavior amplifies the high frequency sensor noise from the actual signal of $\tau_c$. ($\tau_c$ is usually de-noised by a low-pass filter beyond the frequency of $\omega_{SEA}$ which makes the closed loop even more unstable.) Therefore, the augmentation controller must lower the crossover frequency in order to achieve a minimum phase margin.

Similar to \cite{HeThomasPaineSentis2019ACC}, the new crossover cannot be placed at the frequency between $\omega_{h \mhyphen e \mslash \alpha}$ and $\omega_{SEA}$ because multiple other crossovers can be easily triggered. Instead, a new crossover can be safely placed at the frequency between $\omega_{h \mhyphen e}$ and $\omega_{h \mhyphen e \mslash \alpha}$ with a fractional-order augmentation controller
% $C_{\alpha} (s)$ expressed as
\begin{equation} \label{Ca}
C_{\alpha}(s) = k_{f} / s ^ {f},
\end{equation}
where $f$ is the fractional order (that is, a non-integer power of s) of $C_{\alpha} (s)$ and $k_f$ is a gain which allows tuning the magnitude of $C_{\alpha}(s)$ in the frequency domain. The fractional-order controller in \eqref{Ca} has its magnitude decreasing $-20\cdot f$ dB per decade and its phase staying at $-90\cdot f$ degrees at all frequencies. Because of the non zero phase shift from the complex stiffness, a positive phase margin $\phi$ can be guaranteed if
\begin{flalign}
0 < 90 f & < \tan^{-1}(c_h) - \phi,  \notag \\
0 < f & < \tan^{-1}(c_h) / 90 - \phi/90,
\end{flalign}
where $\phi$ is chosen in the range of $(0, \, \tan^{-1}(c_h))$. The value of $k_f$ can then be tuned to achieve a crossover in the frequency range from $\omega_{h \mhyphen e}$ to $\omega_{h \mhyphen e \mslash \alpha}$.

As a fractional-order controller, \eqref{Ca} cannot be implemented directly into the control system. However, we can approximate it as the product of many 1st order lag filters,
\begin{flalign} \label{Cas}
C_{\alpha}(s) & = \frac{k_{f}  \hfill}{p_1 ^ {f}} \cdot \prod_{i=1}^{n} \frac{1 + s/ \, z_i \hfill}{1 + s/p_i}, \\
z_i / p_i & = r_{zp}, \quad \: \text{for} \; \; i = 1, \, 2, \, \cdots, \, n \\
p_i / p_{i-1} & = r_{pp}, \quad \text{for} \; \; i = 2, \, 3, \, \cdots, \, n,
\end{flalign}
where $n$ is the number of lag filters and the pole and the zero for each lag filter are $- p_i$ and $- z_i$. We define $r_{zp}$ such that all the lag filters have an equal distance between the pole and the zero, and we define $r_{pp}$ such that there is a constant distance between adjacent lag filters (in log frequency space). The augmentation controller in \eqref{Cas} functions as a fractional-order filter in the frequency range of $[p_1, \; z_n]$ rad/s. The fractional order $f$ can be approximated as $\log(r_{zp}) / \log(r_{pp})$.

\section{Discussion}

Allowing negative (non-passive) linear damping parameters in M3 makes the model non-passive, but a better fitting F-test comparison model.
%In Tab.~\ref{parameter}, the linear damping parameter in M3 was allowed to be non-passive (negative) because the chirp excitation may cause slight voluntary modulation of muscular activity from the subject \cite{WeissHunterKearney1988JB}. 

In Fig.~\ref{time} the flat peaks of $\theta_e$ could be explained by a hysteretic coulomb-friction-like non-linearity of the human. If this was the case, then our frequency domain measurements would be measurements of the describing function of the non-linearity. 
It is not yet clear how this hypothesis would hold up to testing at different amplitudes of the force input, since we did not include such tests in our experimental plan. It would be a complex task to measure the relationship between the amplitude and the hysteretic damping, since increasing the amplitude would potentially increase the stiffness as well. Regardless of the cause, such hysteresis can not be modeled by M1, and we find that the additional $C_h j$ term in M2 and M3 helps to model the hysteresis by creating a non zero phase shift for the human stiffness. Because a similar type of phase shift has also been identified with passive cockroach legs \cite{DudekFull2006JEB}, we suspect that human joint hysteresis is a passive material property resulting from the contracted muscles around the joint.

As for the proposed fractional-order controller, if our frequency domain model was due to hysteretic coulomb-friction-like non-linearity of the human, then we would expect that the $C_h$ value would be a function of the interaction force signal amplitude. Fitting the model to tests performed at some maximal force amplitude, our controller would also be stable for lower amplitudes. The result would be that our controller would perform well below a force threshold, at which point we would need to switch to another controller---perhaps saturating the desired force or employing a backup safety controller \cite{ThomasHeSentis2018ACC}---to avoid having a non-robust controller in situations with high force amplitude.

This study employed a single subject, so we cannot claim that all humans have damping behavior that fits this model, but some do. More subjects would allow us to be more clear about the observed linear relationship between $K_h$ and $C_h$, and to learn the population variance of our model parameters. Fractional-order controllers can be hand tuned for different users without parameter variance estimates.

\section{Conclusion}

Augmentation exoskeletons rely on a human model to determine stability. While ideal force feedback maps passive environments to passive human experiences, force feedback with finite bandwidth will add energy due to the inevitable phase lag. Human damping directly helps system stability by removing this energy. So the more we know about the damping, the more augmentation can safely be achieved. And in this paper we have presented compelling evidence that this damping is better modeled as hysteretic damping than as viscous damping. With this higher quality model of the human, it should be possible to design augmentation controllers with less conservatism and more performance.
We have additionally theorized a fractional-order controller to take maximal advantage of this model.

% \section{Conclusion}

% \addtolength{\textheight}{-2.cm}   % This command serves to balance the column lengths

%%%%%%%%%%%%%%%%%%%%%%%%%%%%%%%%%%%%%%%%%%%%%%%%%%%%%%%%%%%%%%%%%%%%%%%%%%%%%%%%

%%%%%%%%%%%%%%%%%%%%%%%%%%%%%%%%%%%%%%%%%%%%%%%%%%%%%%%%%%%%%%%%%%%%%%%%%%%%%%%%

%%%%%%%%%%%%%%%%%%%%%%%%%%%%%%%%%%%%%%%%%%%%%%%%%%%%%%%%%%%%%%%%%%%%%%%%%%%%%%%%

%%%%%%%%%%%%%%%%%%%%%%%%%%%%%%%%%%%%%%%%%%%%%%%%%%%%%%%%%%%%%%%%%%%%%%%%%%%%%%%%

% \begin{thebibliography}{99}

% \end{thebibliography}
\bibliographystyle{IEEEtran}
\bibliography{main}

\end{document}